\documentclass[10pt,twocolumn,letterpaper]{article}

\usepackage{cvpr}              

\usepackage{graphicx}
\usepackage{amsmath}
\usepackage{amssymb}
\usepackage{booktabs}
\newcommand\circled[1]{\raisebox{.5pt}{\textcircled{\raisebox{-.7pt} {\footnotesize{#1}}}}}

\usepackage[pagebackref,breaklinks,colorlinks]{hyperref}

\usepackage[capitalize]{cleveref}
\crefname{section}{Sec.}{Secs.}
\Crefname{section}{Section}{Sections}
\Crefname{table}{Table}{Tables}
\crefname{table}{Tab.}{Tabs.}

\def\cvprPaperID{1738} 
\def\confName{CVPR}
\def\confYear{2022}

\newcommand{\argmax}{\mathop{\mathrm{argmax}}}
\begin{document}

\title{DisARM: Displacement Aware Relation Module for 3D Detection}

\author{Yao Duan \quad\quad Chenyang Zhu
\quad\quad Yuqing Lan \quad\quad Renjiao Yi \quad\quad Xinwang Liu \quad\quad Kai Xu\thanks{Corresponding author: kevin.kai.xu@gmail.com}\\
National University of Defense Technology
}
\maketitle

\begin{abstract}
   We introduce Displacement Aware Relation Module (DisARM), a novel neural network module for enhancing the performance of 3D object detection in point cloud scenes. The core idea of our method is that contextual information is critical to tell the difference when the instance geometry is incomplete or featureless. We find that relations between proposals provides a good representation to describe the context. However, adopting relations between all the object or patch proposals for detection is inefficient, and an imbalanced combination of local and global relations brings extra noise that could mislead the training. Rather than working with all relations, we found that training with 
   relations only between the most representative ones, or \emph{anchors},
   can significantly boost the detection performance. 
   A good anchor should be semantic-aware with no ambiguity and independent with other anchors as well.
   To find the anchors, we first perform 
   a preliminary relation anchor module with an objectness-aware sampling approach
   and then devise a displacement based module for weighing the relation importance for a better utilization of contextual information. This light-weight relation module leads to significantly higher accuracy of object instance detection when being plugged into the state-of-the-art detectors. Evaluations on the public benchmarks of real-world scenes show that our method achieves the state-of-the-art performance on both SUN RGB-D and ScanNet V2.
\end{abstract}

\section{Introduction}
\label{sec:intro}
Detecting objects directly from the 3D point cloud is challenging yet imperative in many computer vision tasks, such as autonomous navigation, path planning for robotics, as well as some AR applications. The goal of 3D object detection is to localize all valid shapes and recognize their semantic label simultaneously, which puts forward high requirements for understanding the whole input scene. 

With the rapid development of deep learning and the increasing scale of the online 3D dataset, data-driven methods such as CNN have been widely adopted for object detection. The critical observation of these methods is that the context is as important as the object itself for accurate detection. 
However, the extra information provided by 3D brings noise and irregularity, which makes it more challenging to apply convolution to gather the correct context for detection. 

To avoid irregularity while applying convolution for 3D object detection, the community recently introduced two typical categories of methods. \cite{zhou2018voxelnet, yan2018second, lang2019pointpillars} are trying to project the raw point cloud onto aligned structures such as voxel grids which can apply 3D convolution naturally. In an alternative way, \cite{qi2017pointnet} adopted max-pooling to fuse information of an irregular point cloud directly. These methods can achieve good performance while the input scene is complete and clean. However, the real scanned data is usually incomplete and noisy, making it difficult to extract the key information through this intrinsic context fusion approach. 

\begin{figure}[t]
\centering
\includegraphics[width=0.98\columnwidth]{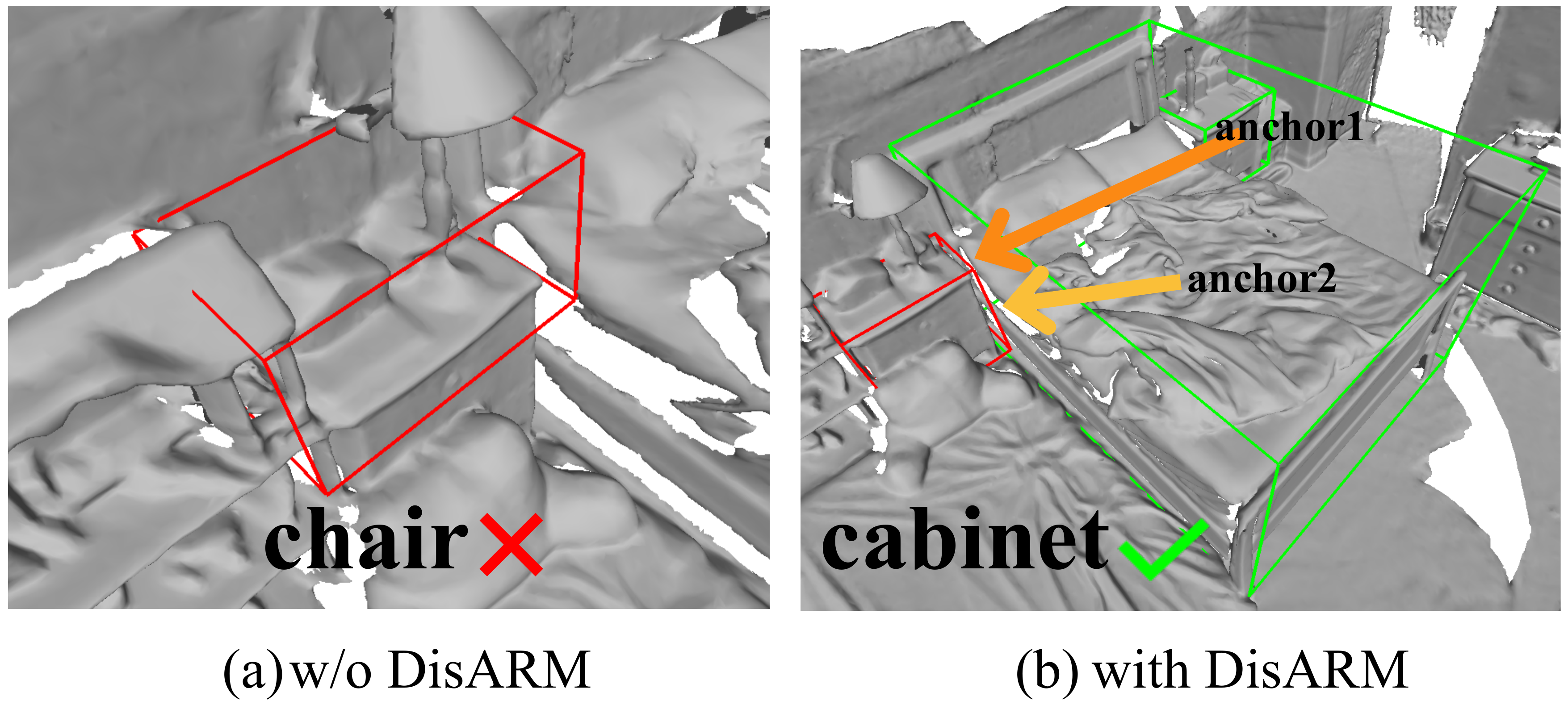} 
\caption{Illustration of the importance of displacement weighted context information for 3D object detection. (a) It is easy to mistake the cabinet as a chair when the point cloud is incomplete and featureless. (b) The network can recognize and locate the cabinet easily with the help of DisARM which provides the surrounding environment information marked as anchors. The colors of arrows indicate the different weights from anchors' displacement.}
\label{fig:teaser}
\end{figure}

To further release the power of context, some methods try to adopt the context explicitly for object detection. Building a relation graph between objects is a natural way to utilize the context. \cite{wald2020learning} leverage inference on scene graphs to enhance 3D scene understanding. However, it requires additional supervision for regression of a correct scene graph. Some methods intend to utilize all the possible relations among the scene to avoid this extra labeling labor. \cite{xie2020mlcvnet} introduces a multi-level framework to fuse all the local and global neighborhoods for 3D object detection. Even a hierarchical architecture is proposed to maintain the context, considering all the relations is still redundant. Furthermore, most methods that adopt context explicitly have their customized network architecture, making it difficult to enhance existing detection methods.

\begin{figure*}[ht]
\centering
\includegraphics[width=0.95\textwidth]{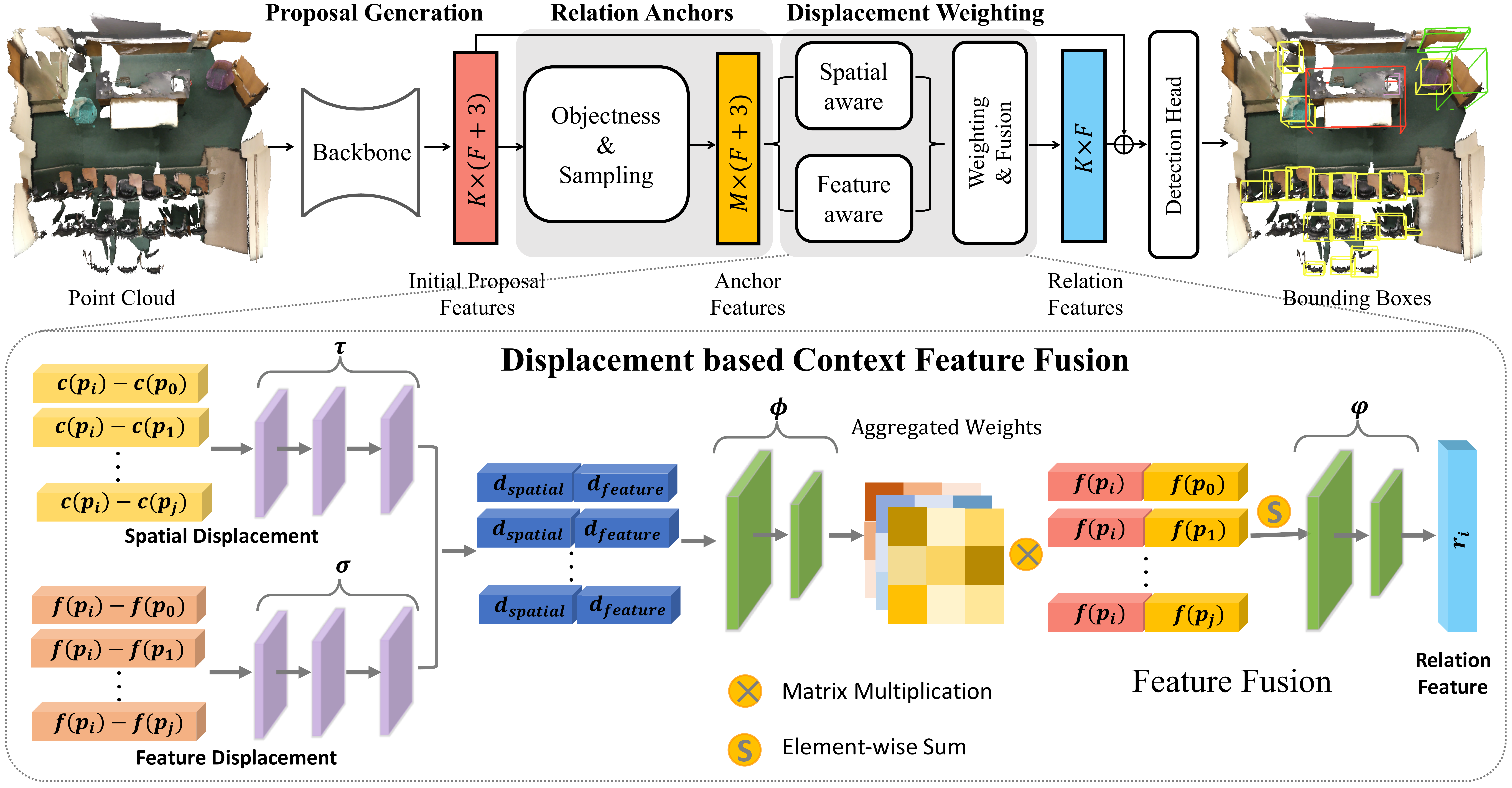} 
\caption{\textbf{DisARM network architecture.} Taking $K$ proposals generated by backbone network as input, we first sample $M$ relation anchors with rich information of the scene's layout. For each of proposals, we get the weights relative to the anchors by considering spatial-aware and feature-aware displacement
. At last, the relation feature is obtained by fusing the weighted proposal-anchor pair features. Note that there is a skip connection operation of relation feature and proposal feature for final detection. In the bottom of figure, $ c(p_{i}) $, $c(p_{j})$ and $f(p_{i})$, $f(p_{j})$ indicate locations and features of proposals and anchors respectively ; $\tau$, $\theta$, $\phi$ and $\varphi$ are the functions consist of MLPs. 
}
\label{fig:pipeline_new}
\end{figure*}

We believe that context fusion is critical for 3D understanding, which can improve object detection performance. 
We introduce a novel neural network module named Displacement Aware Relation Module (DisARM). It can be easily assembled with most existed object detection method and achieves state-of-the-art performance on existing benchmarks. 
The key idea is that context should not only be a structure for information fusion. The relation itself is also a critical feature for 3D understanding. Unlike some previous methods, we try to encode the most critical relations explicitly for potential proposal to allow richer information to be included during the training. 

To avoid the redundant relation features that mislead the training and extract the information that matters, we select and collect the most critical context from two aspects. First, we introduce a relation anchor module, which only sample the most representative and informative proposals as anchors through an objectness-aware \textit{Furthest Point Sampling (FPS)} on feature space. The insight of this design is that the relation anchors for context encoding should distribute uniformly over the feature space while being complete and clean. Our experiments demonstrate that adopting these relation anchors instead of the whole set of relations for context fusion is more efficient and accurate. To maximize the utilization of the proposed relation anchors, we introduce a dynamic weighing mechanism depends on spatial and feature displacement. The key insight here is that the importance of each anchor should be variant regarding recognizing different objects. Due to the object placement usually go with some specific organization pattern for indoor scenes, the importance should depend on the spatial layout and semantic relations between the object and the anchors. In summary, the contributions of this paper include:

\begin{itemize}
\item We propose a portable neural network module that can be assembled with most existing 3D object detection methods to further improve the detection performance, which can be easily implemented as a plug-in for widely used object detection toolbox like MMdetection3D~\cite{mmdet3d2020}. 
\item We introduce a method describing 3D context as a set of weighted representative anchors. This method can effectively extract valid information from the redundant relations in a complex scene.
\item Our method is simple but effective, which achieves 
\textbf{state-of-the-art} performances on ScanNet V2 and \textbf{state-of-the-art} performance with mAP@0.25 on SUN RGB-D.
\end{itemize}


\section{Related Work}
\label{sec:relatedwork}
\subsection{3D object detection on point clouds} 

Object detection from 3D point clouds is challenging due to the irregular, sparse and orderless characteristics of 3D points. Earlier attempts usually relied on projections onto regular grids such as multi-view images \cite{chen2017multi} and voxel grids \cite{zhou2018voxelnet, yan2018second, lang2019pointpillars, hou20193d, shi2020points}, or based on the candidates from RGB-driven 2D proposal generation \cite{qi2018frustum, lahoud20172d} or segmentation hypotheses \cite{kim2013accurate}, where the existing 2D object detection or segmentation methods based on regular image coordinates can be effortlessly adapted. Other approaches also studied how to exploit discriminative \cite{li2015database, nan2012search} or generative shape templates \cite{yi2019gspn}, and high-order contextual potentials to regularize the proposal objectness \cite{lin2013holistic}, or used sliding shapes \cite{song2016deep, song2014sliding}, or clouds of oriented gradients \cite{ren2016three}.

Recent point based detection methods directly process point clouds for 3D object detection. A core task of these methods is to compute object features from the irregularly and sparsely distributed points. All existing methods first assign a group of points to each object candidate and then compute object features from each point group. 
Point R-CNN \cite{shi2019pointrcnn} directly computes 3D box proposals, where the points within this 3D box are used for object feature extraction. PV-RCNN \cite{shi2020pv} leverages the voxel representation to complement the point-based representation in Point R-CNN \cite{shi2019pointrcnn} for 3D object detection and achieves better performance. VoteNet \cite{qi2019deep} groups ponts according to their voted centers and extract object features from grouped points by the PointNet. Some follow-up works further imporve the point group generation procedure \cite{zhang2020h3dnet} or the object box localization and recognition procedure \cite{chen2020hierarchical}.

\subsection{Relation information in 3D object detection}

Contextual information has been demonstrated to be helpful in variety of computer vision tasks, including 2D object detection \cite{hu2018relation, yu2016role}, 
point cloud semantic segmentation \cite{engelmann2017exploring, ye20183d} and 3D scene understanding \cite{zhang2014panocontext, zhang2017deepcontext}. Moreover, the relationships between objects can be treated as a kind of special contextual information which also help the network to improve the performance on computer vision tasks. 

A line of works\cite{zhao2018triangle, wang2021learning} incorporate graphs structures to describe the relationships or exploit the graph convolution networks for relation feature learning. \cite{huang2016structure} defines five types of relations for modeling the graph structure of furniture in indoor scenes, which, however, is time-consuming for relations like facing. \cite{feng2020relation} uses the 3D object-object relation graph structure to explicitly model pairwise relationship information but needs extra supervision. 3DSSG \cite{wald2020learning} defines a rich set of relationships and generates a graph to describe the objects in the scene as well as their relationships which heavily depends on the ground-truth of instance segmentation. HGNet \cite{chen2020hierarchical} leverages graph convolution network to promote performance by reasoning on proposals, while it might not be so useful if the features for detecting an object had been adequately learned.

Another line of works capture the relation information by incorporating features of objects in a variety of ways by neural networks which usually are accompanied by attention mechanisms.  
SRN \cite{duan2019structural} reasons about the local dependencies of regions with considering the inner interactions of point sets by modeling their geometrical and locational relations, which is not suitable for large indoor scenes understanding. 
MLCVNet \cite{xie2020mlcvnet} addresses the 3D object detection task by incorporating into VoteNet multi-level contextual information with the self-attention mechanism and multi-scale feature fusion by considering relations of all objects which results in information redundancy. While MonoPair~\cite{chen2020monopair} learns the pair-wise relationship by only considering the spatial information.


\section{DisARM module}
\label{sec:method}
 \subsection{Overview}
 Some cognitive psychology theories\cite{hu2018relation,engelmann2017exploring,zhang2014panocontext,zhang2017deepcontext} suggest that context can enhance the perception ability for detection. This paper proposes a portable network module, say DisARM, to utilize the 3D context effectively, which can be easily assembled with existing object detection methods to enhance performance.
 
 In our case, we argue that useful contextual information for detection in indoor scenes needs to meet two criteria: it can reflect the intra-relationship between objects and implicitly represent the layout of the entire scene.
 Therefore, a 2-way network framework is proposed to extract the context effectively. As demonstrated in Figure~\ref{fig:pipeline_new}, the former module of DisARM samples the relation anchors between the learned deep feature of each potential object proposal and the following module takes the relative displacement of each proposal between anchors to encode the scene layout.
 More specifically, the core of former module is locating the most representative and informative proposals for relation feature construction. We denote these selected proposals as anchors (see Section~\ref{sec:anchors}). The following module calculates the weights for each anchor through the analysis of spatial and feature displacement (see Section~\ref{sec:displacement}). Our experiments demonstrate that the proposed framework can extract the context for 3D object detection effectively and improve the performance significantly over some state-of-the-art alternatives. 

 \subsection{Relation anchors}\label{sec:anchors}
 \paragraph{Initial proposals} Our DisARM requires initial object proposals $\mathcal{P}=\{p_0,p_1,...,p_n\}$ to boost the relation analysis. VoteNet \cite{qi2019deep} is a widely used 3D detection network that can provide good object proposals. However, it lacks the consideration of the relationships between objects and surroundings. We adopt VoteNet \cite{qi2019deep} as the backbone to produce the input object proposals for DisARM. Note that DisARM can also aggregate with some other detection methods \cite{cheng2021back, liu2021group, qi2020imvotenet}. The evaluation are demonstrated in Table~\ref{table:scannet_sota}.

 Each proposal $p_i$ is represented with its center point.
 The feature encoder network has several multi-layer perception (MLP) layers and feature propagation layers with skip connections. The output feature $f(p_i)$ is an $F$ dimension vector, which is the aggregation of the learned deep feature of each vote that support the proposal $p_i$.

 \begin{figure}[t]
    \centering
    \includegraphics[width=0.4\textwidth]{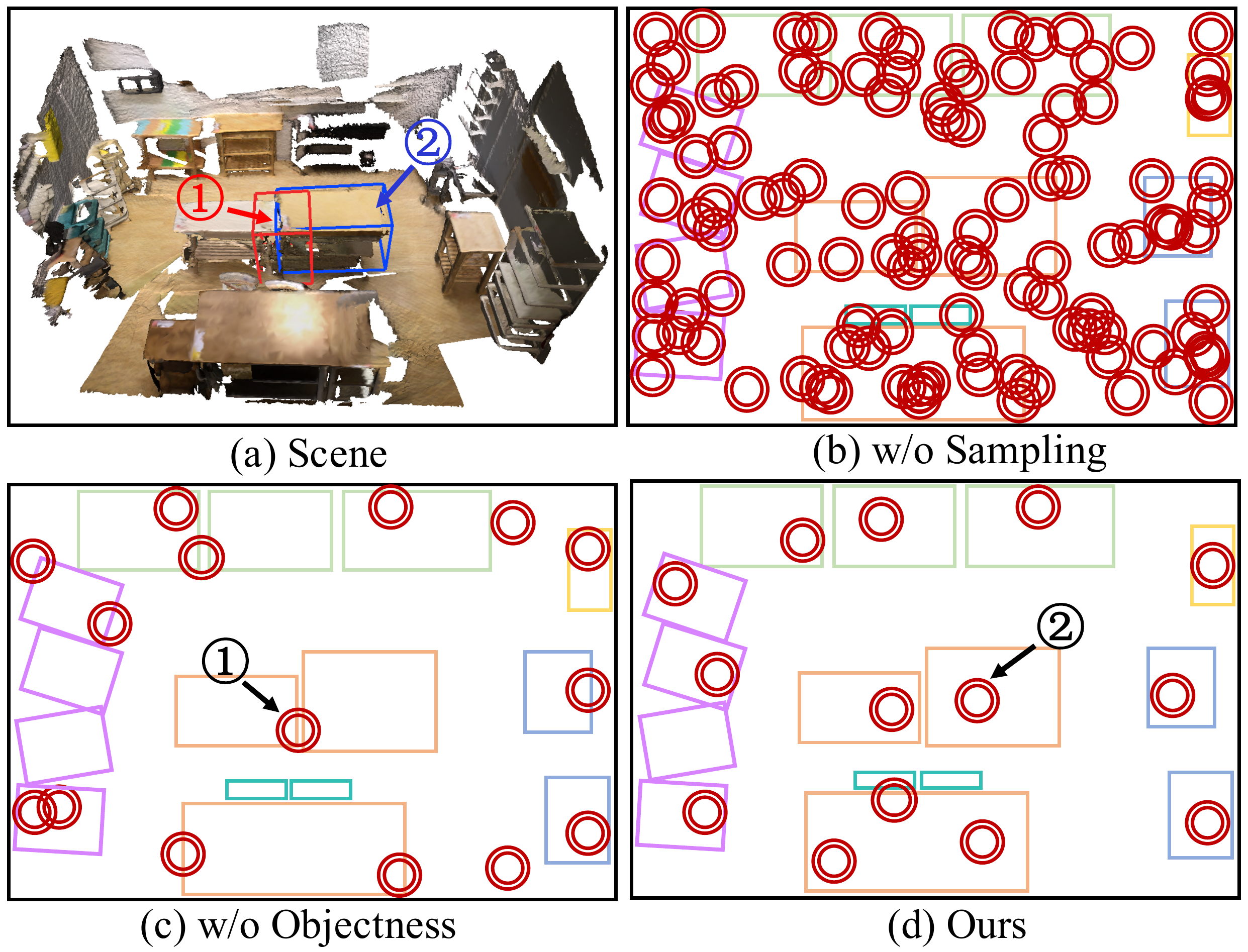} 
    \caption{Anchors. In order to show the results more intuitively, we draw the anchors in red circles and objects in rectangles. Anchors without sampling in Figure(b) are redundant. 
But some anchors sampled by FPS (Figure(c)) are incomplete and invalid, such as anchor \circled{1} which contains parts of the tables. Therefore, we sample on the high objectness anchors. Best view in Figure(a).}
    \label{fig:anchors}
    \end{figure}

 \paragraph{Proposal objectness} As shown in Figure~\ref{fig:anchors}, the whole set of $\mathcal{P}$ is somehow redundant and contains massive incomplete and invalid proposals. Considering all the possible relations in a scene to formulate a context feature is ineffective and may introduce too much noisy information. Therefore, the key to designing a mechanism for utilizing these relations effectively is locating the most representative and informative ones. Figure~\ref{fig:anchors} demonstrates only few proposals given by the backbone are complete. We introduce the concept of objectness to filter the incomplete and noisy ones.

 Given a proposal $p_i$ and its corresponded feature $f(p_i)$, we denote its objectness as $o(p_i)$. The network module to calculate the objectness is a simple MPL network with fully connected layers, sigmoid activation and batch normalization. Since most datasets only label all the valid objects $\mathcal{P}_\text{gt}$ in a scene, we define the objectness loss as below

 \begin{eqnarray}
    \text{loss}_\text{obj}=\| o(p_i)-\chi_{\mathcal{P}_\text{gt}}(p_i)\|\\
    p_i\in\mathcal{P}_\text{gt} \iff \exists p\in\mathcal{P}_\text{gt}\rightarrow\text{IoU}(p_i, p)>0.25
\end{eqnarray}

where $\chi_{\mathcal{P}_\text{gt}}(p_i)$ is an indicator function. As demonstrated in Figure~\ref{fig:anchors}, $o(p_i)$ can indicate the completeness of a given proposal which is critical for locating the proposal anchors.

\paragraph{Anchor sampling} Even we only focus on complete proposals, the aggregation of $\mathcal{P}$ appears to be evident. Previous works such as \textit{KPS} in \cite{liu2021group} which only focus on high objectness proposals will still introduce redundant information.
We find that conducting \textit{Furthest Point Sampling (FPS)} on $\mathcal{P}$ with the assistance of objectness evaluation can help us locate the most representative proposal anchors. 

In details, a proposal $p_0$ with the highest objectness score is first sampled in $\mathcal{P}_{\text{anchor}}^{(0)}$. The next sampling would be processed as below,
\begin{equation}\label{eq:sample}
    \mathcal{P}_{\text{anchor}}^{(k+1)} = \{\mathcal{P}_{\text{anchor}}^{(k)}, \argmax_{p_i\in\mathcal{P}} \sum_{p_j\in\mathcal{P}_{\text{anchor}}^{(k)}}o(p_i)\|f(p_i)-f(p_j)\|\}
\end{equation}
Eq.~\ref{eq:sample} indicate the metric adopting in our Furthest Point Sampling (FPS) is upon the feature space $f(\cdot )$ weighted by the objectness score $o(\cdot )$. Then the farthest proposal $p_i$ to the already-chosen proposal set $\mathcal{P}_{\text{anchor}}^{(k)}$ is iteratively selected until the number of chosen proposals meets the candidate budget $ M $, which is 15 for all our evaluations. Though it is simple, the finally selected $ M $ proposal anchors are representative and distributed uniformly in the whole scene.

\begin{figure}[t]
    \centering
    \includegraphics[width=0.9\columnwidth]{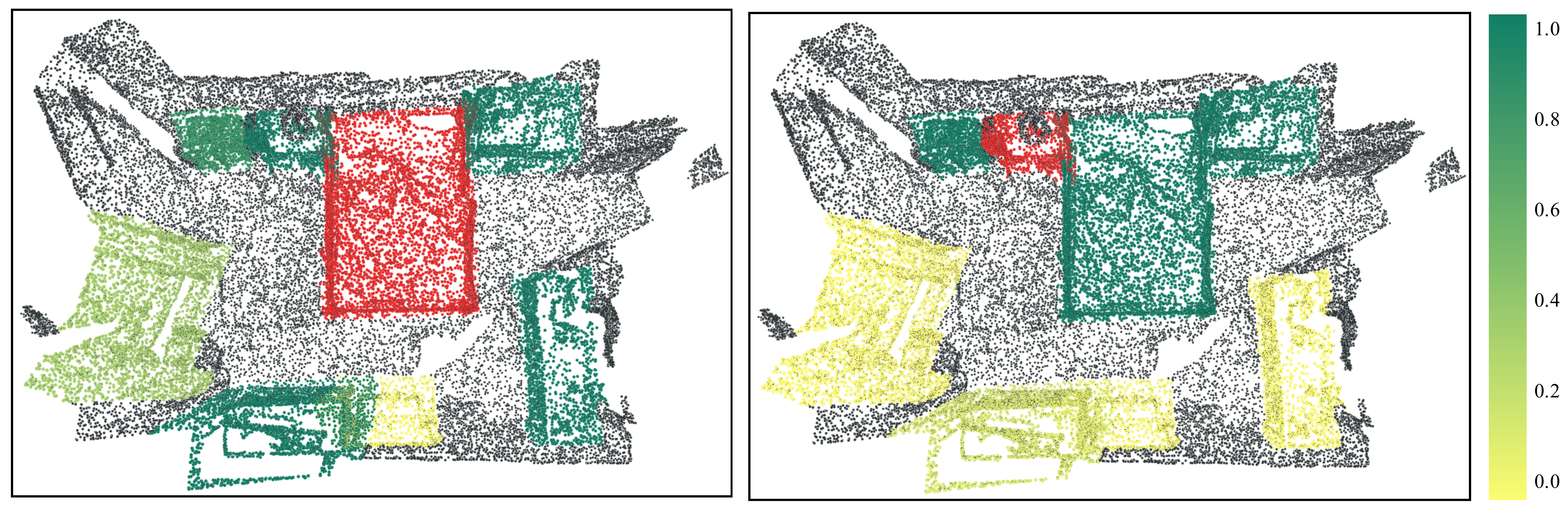} 
    \caption{Displacement weights. We show the proposal in red points and anchors with different colors correspond to different weights.  Bed 
has different perception of beside cupboards and shelves which is effected by spatial displacement. Beside cupboard in the same scene is more interested in another beside cupboard and cabinet which is determined by feature displacement.}
    \label{fig:displacement}
\end{figure}

\subsection{Displacement based context feature fusion}\label{sec:displacement}
\paragraph{Spatial displacement} The proposal anchors $\mathcal{P}_{\text{anchor}}$ can effectively describe the context of the whole input scene. However, they should not contribute equally for detection of different objects as demonstrated in Fig~\ref{fig:displacement}. Adopting appropriate anchors is critical to utilize context in detection. Inspired by \cite{xu_sig14}, spatial layout patterns can effectively describe the representative substructures in an indoor scene. Therefore, we think the context for detection should be weighted by layout-aware spatial displacement as well.

We argue that an object has different perceptions towards different proposal anchors regarding with different spatial displacement. For example, cabinets are usually placed next to bed and chairs are most commonly placed in front of a desk or table. These patterns can be reflected by spatial displacement among proposal-anchor pairs. Thus, we regard the importance of different displacement around proposals as displacement weights which encourages networks to pay different levels of attention. For details, given the target proposal $p_i$ with location $c(p_i)$ and a proposal anchor $p_j$ with location $c(p_j)$, the spatial displacement weight between them is formulated as $d_{\text{spatial}}(p_i,p_j)=\tau(c(p_i) - c(p_j))$, where $\tau$ is a perception function given by an MLP network.

\paragraph{Feature displacement} Similar with the spatial displacement, the feature displacement $f(p_i)-f(p_j)$ given by the target proposal $p_i$ and proposal anchor $p_j$ should be also considered while measuring the importance of the proposal-anchor pair. The insight here is, layout patterns are sometimes semantic-aware. For example, the existence of a bathtub would always indicate a washbasin in the scene. This characteristic can be reflected by the pre-encoded features $f(p_i)$ and $f(p_j)$ since objects with similar semantic label would also be close on the feature space and vice versa. Therefore, given the target proposal $p_i$ and a proposal anchor $p_j$, the feature displacement weight between them is formulated as $d_{\text{feature}}(p_i,p_j)=\sigma(f(p_i) - f(p_j))$, where $\sigma$ is a perception function given by an MLP network.

\paragraph{Aggregated weights} We concatenate spatial displacement weight $d_{\text{spatial}}(p_i,p_j)$ feature displacement weight $d_{\text{feature}}(p_i,p_j)$ together to fuse the perceived information before put them into an MLP network as shown in Figure \ref{fig:pipeline_new}. We can get final aggregated weights as below,

\begin{equation}
\label{eq:weights}
w(p_i,p_j) =  \text{tanh}(\phi (d_{\text{spatial}}(p_i,p_j); d_{\text{feature}}(p_i,p_j)))
\end{equation}

where $\phi$ is a perception function enabled by several MLP layers. To further normalize the weights between $p_i$ and all the anchors in $\mathcal{P}_{\text{anchor}}$, we adopt softmax function in the end.

\begin{equation}
    \label{eq:normalize}
    w(p_i,p_j) = \frac{w(p_i,p_j)}{\sum_{p_k\in\mathcal{P}_{\text{anchor}}}w(p_i,p_k)} 
\end{equation}

Finally, We formulate the fused relation feature $r_i$ of an object proposal $p_i$ for detection as below,
 \begin{equation}
     r_i = \varphi(\sum_{p_j\in \mathcal{P}_{\text{anchor}}}w(p_i, p_j)\cdot [f(p_i);f(p_j)])
 \end{equation}
However, it is obvious that training $f(\cdot)$, $w(\cdot)$ and finding the optimal $\mathcal{P}_{\text{anchor}}$ is highly correlated which makes it a challenging optimization problem. We propose a 3-stage framework to find the optimal $r_i$. At the warm-up stage, $w(p_i, p_j)$ is set inactive and the proposed module focus on locating the optimal $\mathcal{P}_{\text{anchor}}$ and training $f(p_i)$. The insight of this design is $w(p_i, p_j)$ would only be functional while the network can already extract some reasonable proposal anchors. At the next stage, we freeze $\mathcal{P}_{\text{anchor}}$ and $f(p_i)$ to optimize $w(p_i, p_j)$. This design would fully utilize layout information extracted from the scene to measure the anchor importance. After these two stages, $w(p_i, p_j)$, $\mathcal{P}_{\text{anchor}}$ and $f(p_i)$ are fine-tuned together to achieve the final optimal.


\section{Experiments}
\label{sec:exp}

\begin{table}[t]
    \centering
    \setlength{\tabcolsep}{0.72mm}
    \renewcommand\arraystretch{1.2}{
    \begin{tabular}{lll}
    \hline
    Method                                       & ${\text{mAP}} _{25}$   & ${\text{mAP}}_{50}$     \\ 
    \hline \hline
    HGNet~\cite{chen2020hierarchical}            & 61.3        & 34.4        \\
    GSPN~\cite{yi2019gspn}                       & 62.8        & 34.8        \\
    Pointformer+~\cite{pan20213d}                & 64.1        & -           \\
    3D-MPA~\cite{engelmann20203d}                & 64.2        & 49.2        \\
    MLCVNet~\cite{xie2020mlcvnet}                & 64.7        & 42.1        \\
    \hline
    VoteNet~\cite{qi2019deep}                    & 58.6        & 33.5        \\ 
    VoteNet*   									 & 63.8        & 44.2            \\
    VoteNet*+DisARM                              & 66.1 \textcolor{green}{$ \uparrow $}        & 49.7 \textcolor{green}{$\uparrow$}        \\
	\hline    
	BRNet~\cite{cheng2021back}                   & 66.1        & 50.9        \\
    BRNet+DisARM                                 & 66.7 \textcolor{green}{$\uparrow$}         & 52.3 \textcolor{green}{$\uparrow$}            \\ 
	\hline    
	H3DNet*~\cite{zhang2020h3dnet}                & 66.4        & 48.0        \\   
    H3DNet*+DisARM                                & 66.8 \textcolor{green}{$\uparrow$}              & 48.8 \textcolor{green}{$\uparrow$}             \\  
    \hline
    GroupFree3D*(L6, O256)~\cite{liu2021group}             & 66.3        & 47.8 \\ 
    GroupFree3D*(L12, O256)~\cite{liu2021group}            & 66.6     & 48.2 \\
    GroupFree3D*(w2$\times$, L12, O512)~\cite{liu2021group}            & 68.2     & 52.6 \\
    GroupFree3D*(L6, O256)+DisARM                          & 67.0 \textcolor{green}{$\uparrow$}         & 50.7 \textcolor{green}{$\uparrow$}        \\
    GroupFree3D*(L12, O256)+DisARM                          & 67.2 \textcolor{green}{$\uparrow$}         & 52.5 \textcolor{green}{$\uparrow$}        \\
    GroupFree3D*(w2$\times$, L12, O512)+DisARM                          & \textbf{69.3} \textcolor{green}{$\uparrow$}         & \textbf{53.6} \textcolor{green}{$\uparrow$}        \\
    \hline  
    \end{tabular}
    \caption{3D object detection results on ScanNet V2 dataset. \textbf{Notations:} We report the detection performance using mean Average Precision (mAP) at IoU thresholds of 0.25 and 0.5, denoted as ${\text{mAP}}_{25}$ and ${\text{mAP}}_{50}$. Pointformer+ indicates the VoteNet equipped with Pointformer and * denotes that the model is implemented on MMDetection3D. We denote VoteNet*+DisARM, BRNet+DisARM and GroupFree3D*+DisARM as enhanced versions with our method respectively, $\uparrow$ indicates the performance is improved with the equipment of DisARM.}
    \label{table:scannet_sota}}
    \end{table}

\begin{figure}[t]
    \centering
    \includegraphics[width=0.45\textwidth]{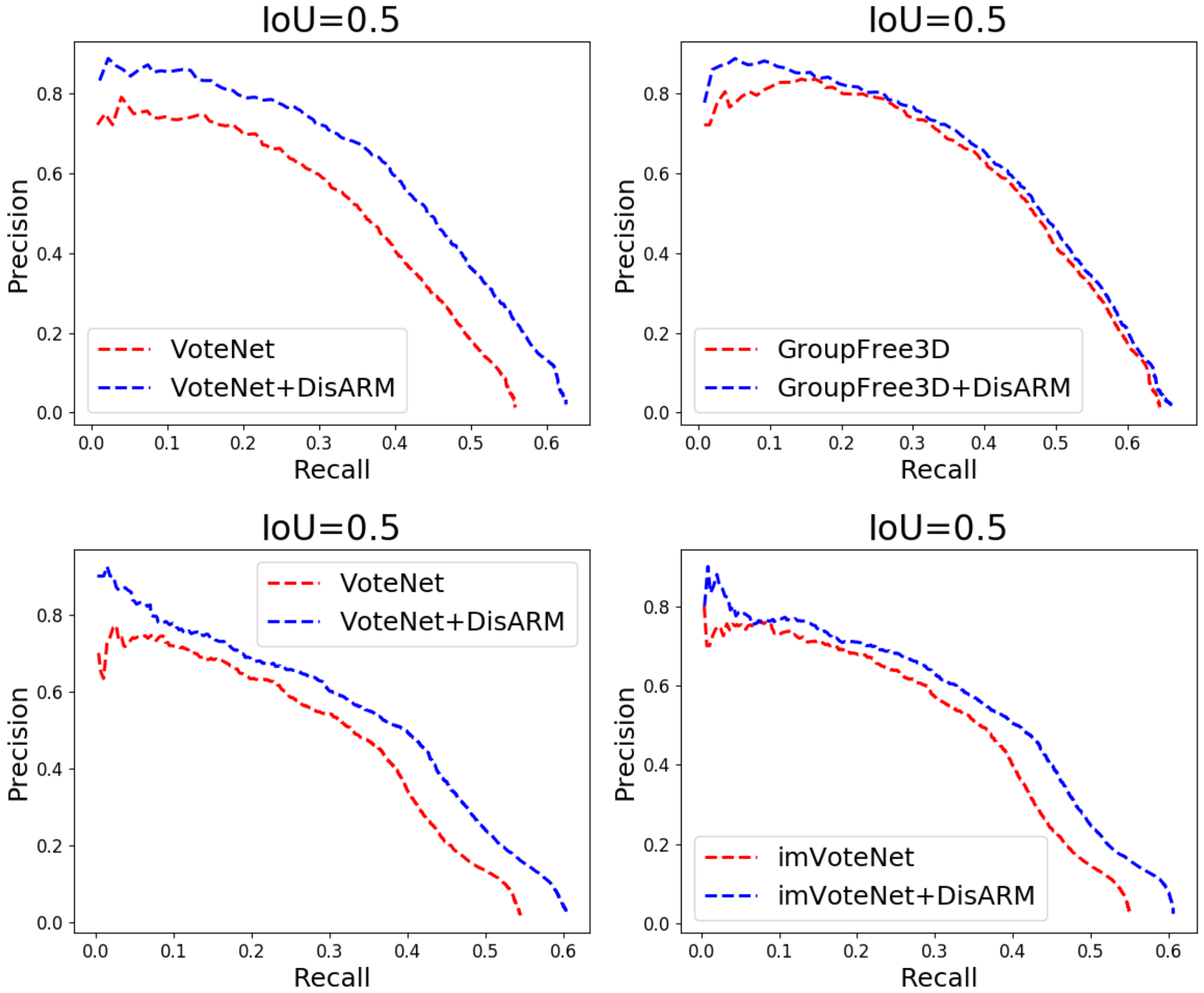} 
    \caption{Precision-Recall Curves of different backbones equipped with our DisARM on mAP@0.5. We show the results on ScanNet V2 dataset in the first row and the results on SUN RGB-D dataset in the second row.}
    \label{fig:pr}
\end{figure}

\begin{figure*}[!t]
\centering
\includegraphics[width=2\columnwidth]{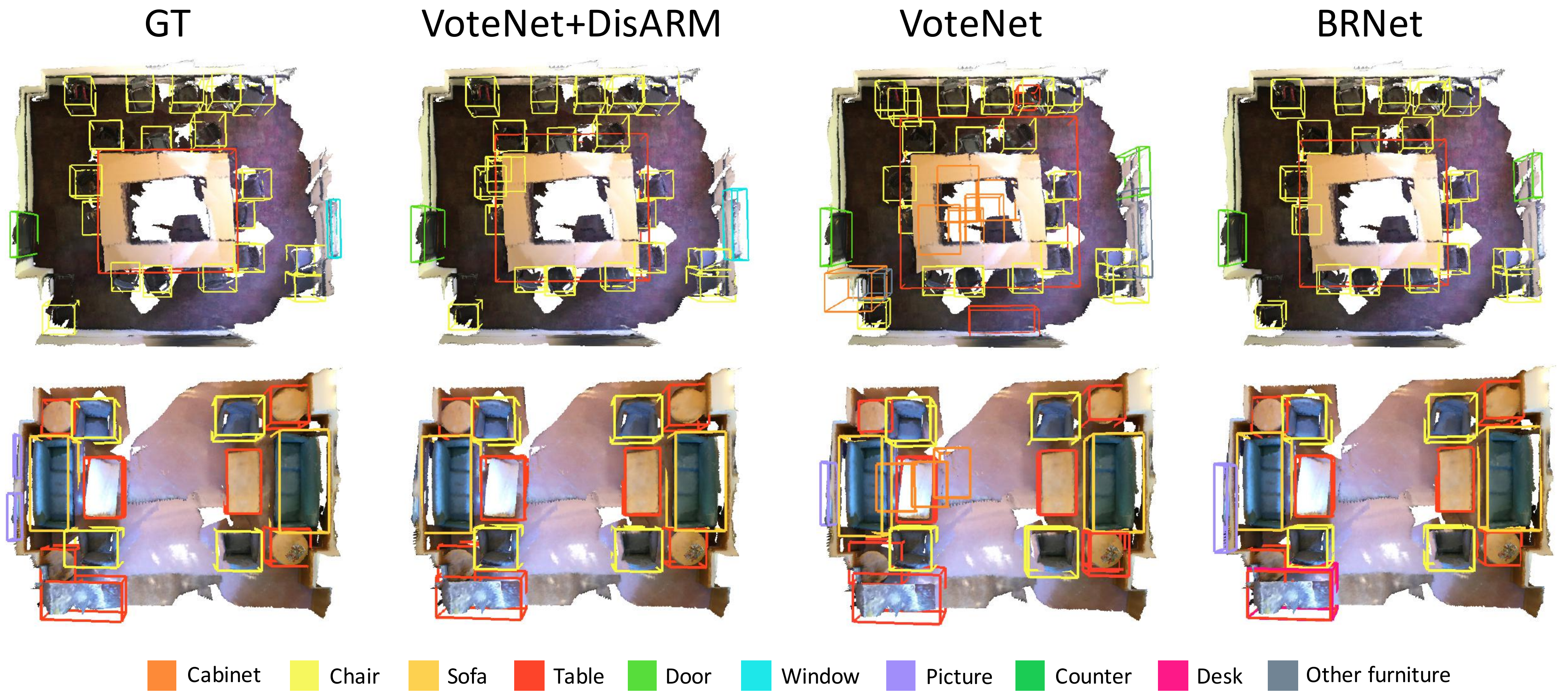}
\caption{Qualitative results on ScanNet V2 dataset. We denote VoteNet+DisARM as applying our method to VoteNet. The first column is ground truth and the rest columns are detections of different methods. Best viewed on screen.}
\label{fig:results}
\end{figure*}

\begin{table*}[!t]
    \centering
    \setlength{\tabcolsep}{1.4mm}
    \renewcommand\arraystretch{1.1}{
    \begin{tabular}{c|cccccccccc|c}
    \hline
    \textbf{}    & bathtub  & bed  & bookshelf  & chair & desk &dresser  & nightstand & sofa & table & toilet & mAP\\  \hline

    VoteNet~\cite{qi2019deep}       & 74.4  & 83.0 & 28.8 & 75.3 & 22.0 & 29.8 & 62.2 & 64.0 & 47.3 & 90.1 & 57.7  \\  
  
    BRNet~\cite{cheng2021back}  & 76.2  & 86.9 & 29.7 & 77.4 & 29.6 & 35.9 & 65.9 & 66.4 & 51.8 & 91.3 & 61.1  \\

    GroupFree3D~\cite{liu2021group} & \textbf{80.0} & \textbf{87.8} & 32.5 & 79.4 & 32.6 & 36.0 & 66.7 & 70.0 & \textbf{53.8} & 91.1 & 63.0\\   

    imVoteNet~\cite{qi2020imvotenet} & 75.9 & 87.6 & 41.3 & 76.7 & 28.7 & \textbf{41.4} & \textbf{69.9} & 70.7 & 51.1 & 90.5 & 63.4 \\  \hline

    VoteNet+DisARM  & 76.7 & 86.2 & 35.4 & 78.4 & 31.0 & 34.6 & 66.3 & 68.1 & 51.2 & 86.9 & 61.5 \\  
    imVoteNet+DisARM  & 79.9 & 87.5 & \textbf{43.7} & \textbf{80.7} & \textbf{33.3} & 39.8 & 69.5 & \textbf{74.1} & 52.7 & \textbf{91.6} & \textbf{65.3}\\  
    \hline
    \end{tabular}}
    \caption{3D object detection results  on SUN RGD-D val dataset with mAP@0.25. \textbf{Notations:} * denotes that the model is implemented on MMDetection3D. VoteNet*+DisARM and imVoteNet+DisARM indicate applying our method to the 3D object detectors respectively.}
    \label{table:sunrgb_sota}
\end{table*}

\begin{table}[t]
\centering
\setlength{\tabcolsep}{1.5mm}
\renewcommand\arraystretch{1.3}{
\begin{tabular}{lcc}
    \hline
    Settings                                        & mAP@0.25      & mAP@0.5     \\ \hline
    \circled{1} Global                  & 63.3               & 47.7        \\
    \circled{2} Local                   & 64.3                & 48.2       \\
    \circled{3} Random               & 65.0               & 48.7 \\ 
    \hline
    \circled{4} D-FPS               & 65.1        & 48.7        \\
    \circled{5} F-FPS                 & 65.3         & \textbf{49.7}       \\
    \circled{6} D-FPS+F-FPS                 & 65.0          & 48.8        \\ 
    \circled{7} F-FPS+D-FPS                 & 65.3         & 48.4        \\ 
    \hline
    \circled{8} K-means                 & 65.2         & 48.2        \\ 
    \circled{9} K-means+D-FPS                 & 65.0          & 48.7        \\ 
    \circled{10} K-means+F-FPS                 & 64.4          & 48.3        \\ 
    \hline
    \circled{11} Ours                 & \textbf{66.1}          & \textbf{49.7}        \\ 
    \hline

\end{tabular}
\caption{Ablation studies of sampling relation anchors strategies. Note that the experiments \circled{1} to \circled{3} indicate selecting relation anchors by taking all proposals (Global), nearest 15 proposals (Local) and  random 15 proposals (Random) as anchors. Experiments \circled{4} to \circled{7} use different combinations of FPS on distances(D-FPS) and features(F-FPS). Experiments \circled{8} to \circled{10} sample anchors on clusters generated by K-means. Experiment \circled{11} conducts FPS on the anchors filtered by objectness score (OS). }
\label{table:anchors}}
\end{table}

\begin{table}[t]
\centering
\setlength{\tabcolsep}{0.7mm}
\renewcommand\arraystretch{1.3}{
\begin{tabular}{c|ccccc|c}
    \hline
    DCFF            & window      & desk    & showr      & toil          & sink        & ${\text{mAP}}_{50}$     \\ \hline
    w/o              & 22.7                 & 44.4    & 36.8        &  86.4       & 37.5       & 47.1                 \\  
    S-DW          & 26.2                   & 46.3    & 31.4         & 89.7   & 37.3             & 47.9                  \\ 
    F-DW          & 20.2                    & 48.6    & 45.7         & 89.2   & 36.7            & 48.9                  \\ 
    Ours         & \textbf{27.5}     & \textbf{55.1}    & \textbf{49.8}   & \textbf{91.4}   & \textbf{44.5}  & \textbf{49.7}                  \\ 
    \hline

\end{tabular}
\caption{DisARM with different components in displacement based context feature fusion (DCFF). The first row indicates fusing features of proposals and anchors without weighting. We denote the S-DW, F-DW as learning weights by spatial displacement and feature displacement respectively. }
\label{table:dam}}
\end{table}

\begin{table}[t]
    \centering
    \setlength{\tabcolsep}{0.8mm}
    \renewcommand\arraystretch{1.15}{
    \scalebox{0.9}{
    \begin{tabular}{c|c|c|c|c}
        \hline
        Method                                & Model size &  time     &GFLOPs   & mAP@0.5  \\ \hline
        VoteNet*                                  & 11.6MB     & 0.095s  & 5.781  & 44.2 \\
        
        BRNet                                 & 13.2MB      & 0.132s  & 7.97  & 50.9 \\
        GroupFree3D*                             & 113.0MB    & 0.170s & 31.05 & \textbf{52.6}  \\ \hline         
        VoteNet*+DisARM                          & +1MB     & +0.001s  & +0.034 & 49.7  \\ 
        BRNet+DisARM                            & +1MB       & +0.008s  & +0.034 & 52.3 \\ \hline
    \end{tabular}
    }
    \caption{Comparison of model size for different methods. * denotes the model implemented in MMDetection3D \cite{mmdet3d2020}. GroupFree3D~\cite{liu2021group} reported here is with best-performance setting. 
}
    \label{table:modelsize}}
    \end{table} 
    
\begin{figure*}[t]
\centering
\includegraphics[width=1.9\columnwidth]{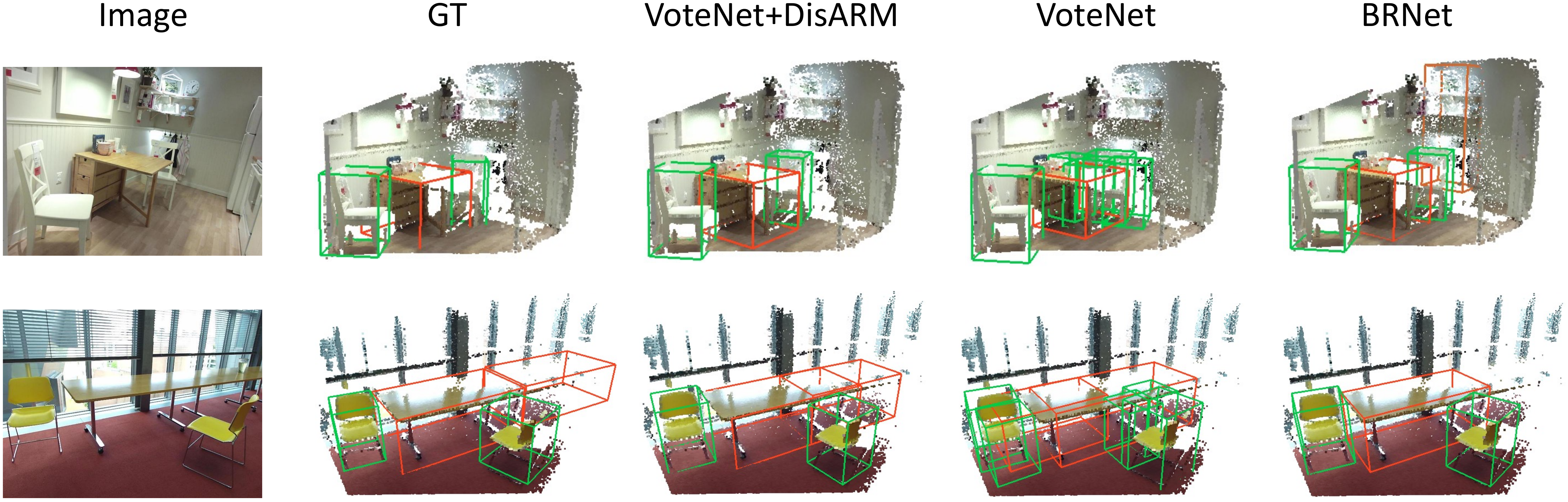} 
\caption{Qualitative results on SUN RGB-D dataset. We denote VoteNet+DisARM as applying our method to VoteNet. The first column is ground truth and the rest columns are detections of different methods. Best viewed on screen.}
\label{fig:results_sunrgbd}
\end{figure*}

As our method can be applied to several backbones, we describe the implementation based on VoteNet~\cite{qi2019deep} in brief. More details of other backbones in the Supplementary. In our DisARM, we take the 256 output proposals of VoteNet~\cite{qi2019deep} with 128-dimension features as input.  And then we use a MLP network to predict objectness and subsample 128 candidate anchors according the scores. The MLP is realized
with FC output sizes of 64, 32, 32, 1, where the final objectness scores are obtained by the output of last layer followed by sigmoid function. The function $\tau$ for spatial displacement has 3 layers of 8, 16, 32 hidden dimensions and function $\sigma$ for feature displacement has 2 layers of 64, 32 hidden dimensions. The MLP hidden dimensions are 32, 1 of function $\phi$ for aggregated weights. The relation encoder $\varphi$ for relation feature $r_{i}$ has 4 layers of 256, 128, 128, 128 hidden dimensions.

We evaluate our method on two widely-used 3D object detection datasets: ScanNet V2 \cite{dai2017scannet} and SUN RGB-D \cite{song2015sun}. Standard data splits in \cite{qi2019deep} are adopted. Our network is end-to-end optimized with the batch size of 8. The initial learning rate is 0.008 and the network is trained for 220 epochs on both datasets. The Cosine Annealing \cite{loshchilov2016sgdr} is adopted as the learning rate schedule. We implement our method on MMDetection3D~\cite{mmdet3d2020} with one NVIDIA TITAN V GPU. The source code is provided in Supplementary. 

\subsection{Comparisons}
In this section, we compare our method with previous state-of-the-arts on ScanNet V2 and SUN RGB-D dataset, such as VoteNet\cite{qi2019deep} and its successors MLCVNet \cite{xie2020mlcvnet}, HGNet \cite{chen2020hierarchical}, H3DNet \cite{zhang2020h3dnet}, BRNet \cite{cheng2021back} and so on.

\noindent \textbf{Quantitative results.} The detection results of ScanNet V2 dataset are shown in Table \ref{table:scannet_sota}. Applying our DisARM to VoteNet~\cite{qi2019deep} achieves 66.1 on mAP@0.25 and 49.7 on mAP@0.5 over the implementation in MMDetection3D \cite{mmdet3d2020}, which is \textbf{7.5} and \textbf{16.2} higher than the performance of VoteNet reported in ~\cite{qi2019deep} and outperforms its original version by 2.3 and 5.5 mAP with MMDetection3D. 

Applying our DisARM to better 3D object detectors like H3DNet~\cite{zhang2020h3dnet}, BRNet~\cite{cheng2021back}, GroupFree3D~\cite{liu2021group}, we obtain 0.4, 0.6, 0.7 improvement on mAP@0.25 and 0.8, 1.4, 2.9 improvement on mAP@0.5 respectively. Furthermore, DisARM applied to GroupFree3D~\cite{liu2021group} with strongest backbone achieves the \textbf{state-of-the-art} performance on mAP@0.5.

The substantially improved results 
indicate that our DisARM can not only benefit basic 3D object detectors but also help improve complex detectors which demonstrates the high effectiveness and generalization of our method. It is noteworthy that
VoteNet*+DisARM outperforms  GroupFree3D* using 12 attention modules on mAP@0.5, which indicates that our method is simple but more effective than those methods with complicated architectures. The results of more improved performance on mAP@0.5 which is a fairly challenging metric show that DisARM helps the backbones to detect the objects more accurately attributing success to our method eliminating ambiguity with the relational context information. 
We also draw the PR curves of different methods equipped with DisARM in Figure \ref{fig:pr}.

As shown in Table. \ref{table:sunrgb_sota}, we compare with previous state-of-the-arts on SUN RGB-D dataset. In the same way, we evaluate our method on VoteNet which outperforms VoteNet on mAP@0.25 by 2.4 and mAP@0.5 by 5.5. In particular, our DisARM applied to imVoteNet~\cite{qi2020imvotenet} achieves \textbf{65.3} on mAP@0.25, which outperforms all previous state-of-the-arts. More quantitative results on ScanNet V2 and SUN RGB-D datasets can be found in Supplementary.

\noindent \textbf{Qualitative results.} In Figure \ref{fig:results} and Figure \ref{fig:results_sunrgbd}, we visualize the representative 3D object detection results from our method and the baseline methods. 
These results demonstrate that applying our method to baseline detector achieves more reliable detection results with more accurate bounding boxes and orientations. Our method also eliminates false positives and discovers more missing objects when compared with the baseline methods. For example, the results in the second row of Figure \ref{fig:results_sunrgbd} show that there are two tables in the scene, and the left one is complete while the right one is missing partially. Our method VoteNet+DisARM can basically detect the right table (red boxes), while all the other methods miss the challenging one. This proves that our method can provide rich and effective context to boost the performance of 3D object detectors. More qualitative visualizations are shown in Supplementary.

\subsection{Ablation Study}

We conduct extensive ablation experiments to analyze the effectiveness of different components of DisARM. All experiments are trained and evaluated on the ScanNet V2 dataset and take VoteNet \cite{qi2019deep} as backbone method. The network is implemented in MMDetection3D~\cite{mmdet3d2020}.

\noindent
\textbf{Strategies of sampling relation anchors.} As shown in Table \ref{table:anchors}, 
applying DisARM to VoteNet using our sample strategy achieves the highest performance. Experiment \circled{1} and experiment \circled{2} shows that both global and local context can not provide effective information which introduce redundant information or limited information. We also find that conduct FPS on proposal features can keep the diversity of anchors which can provide more useful context through experiments \circled{4}\circled{6} and experiments \circled{5}\circled{7}.

Clustering by K-means is a common way to aggregate information. Thus we try to conduct D-FPS and F-FPS on clusters generated by K-means as shown in experiments \circled{8}\circled{9}\circled{10}. Those strategies can not perform best on mAP@0.5 for the reason that the aggregated context loss the key information of objects for accurate detection. We argue that complete objects are representative and informative as anchors. And experiment \circled{11} proves our argument. 

\noindent

\noindent
\textbf{Effects of displacement based context feature fusion.} We evaluate the contribution of displacement weights in DisARM on ScanNet V2 dataset. The quantitative results are shown in the Table. \ref{table:dam}. It is clear that the proposed displacement weights are useful and can distribute accurate weights for contexts from different relation anchors, providing more helpful and robust contexts for better performance. We find that displacement weights are sensitive to the objects usually placed in special space or scenes, such as window, shower curtain, toil and sink. The large improved performance on mAP@0.5 also indicates the effectiveness of our displacement weights design and our DisAM can help the backbones to detect these hard ones more accurately.

\noindent
\textbf{Model size and speed.} The comparison of efficiency is shown in Table \ref{table:modelsize}. For a fair comparison, all experiments are running on the same workstation (a single Titan V GPU) and implemented with MMDetection3D. 
It is obvious that our proposed method is effective with increasing very few training parameters to backbone methods. Note that the model size of BRNet equipped with DisARM is $10\times$ less than GroupFree3D only with little performance dropping. Knowing that the proposed method has significant performance gains than these reference methods listed in Table \ref{table:scannet_sota} and Table \ref{table:sunrgb_sota}, its lightweight model validates that the proposed DisARM is also efficient for 3D object detection.


\section{Conclusion}
\label{sec:conc}
In this paper, we present a simple, lightweight yet effective method for enhancing the performance of 3D object detection. Unlike previous methods detect objects individually or use context information inefficiently, our method samples representative relation anchors and captures the relation information with the contribution of each relation anchor determined by the spatial-aware and feature-aware displacement weights. The proposed method achieves state-of-the-art performance on ScanNet V2 with both metrics and SUN RGB-D in terms of mAP@0.25.

\textbf{Limitation} Our approach is designed for the indoor scenes with some specific organization pattern, and it is not suitable for outdoor scenes with irregular displacement. However, we will explore more relation information for all kinds of scenes in the future.


\section{Acknowledgements}
\label{sec:ack}
This paper is supported in part by the Nation Key Research and Development Program of China (2018AAA0102200), National Nature Science Foundation of China (62132021, 62102435, 62002375, 62002376), NSF of Hunan Province of China (2021RC3071, 2021JJ40696) and NUDT Research Grants (ZK19-30).


{\small
\bibliographystyle{ieee_fullname}
\bibliography{egbib}
}

\end{document}